\documentclass[letterpaper, 10 pt, conference]{ieeeconf}  

\IEEEoverridecommandlockouts                              

\usepackage{algorithm}
\usepackage{algpseudocode}
\usepackage{kotex}
\usepackage{xcolor}
\usepackage{booktabs}
\usepackage{multirow}
\usepackage{xspace}

\usepackage{cite}
\usepackage{amsmath}
\usepackage{amssymb}
\usepackage{amsfonts}
\usepackage[final]{graphicx}
\usepackage{textcomp}
\usepackage{etoolbox}
\usepackage{comment}
\usepackage{hyperref}
\usepackage{cleveref}
\usepackage{flushend}

\title{\LARGE \bf
Observation Space Matters: Benchmark and Optimization Algorithm
}

\author{
Joanne Taery Kim$^{1}$ and Sehoon Ha$^{23}$
\thanks{$^{1}$ Lawrence Livermore National Laboratory, Livermore, CA, 94550, USA}
\thanks{$^{2}$ Georgia Institute of Technology, Atlanta, GA, 30308, USA}
\thanks{$^{3}$ Robotics at Google, Mountain View, CA, 94043, USA }
\thanks{Emails: {\tt\small kim102@llnl.gov, sehoonha@gatech.edu}}
}

\begin{document}

\maketitle
\thispagestyle{empty}
\pagestyle{empty}

\begin{abstract}
Recent advances in deep reinforcement learning (deep RL) enable researchers to solve challenging control problems, from simulated environments to real-world robotic tasks.
However, deep RL algorithms are known to be sensitive to the problem formulation, including observation spaces, action spaces, and reward functions.
There exist numerous choices for observation spaces but they are often designed solely based on prior knowledge due to the lack of established principles.
In this work, we conduct benchmark experiments to verify common design choices for observation spaces, such as Cartesian transformation, binary contact flags, a short history, or global positions.
Then we propose a search algorithm to find the optimal observation spaces, which examines various candidate observation spaces and removes unnecessary observation channels with a Dropout-Permutation test.
We demonstrate that our algorithm significantly improves learning speed compared to manually designed observation spaces.
We also analyze the proposed algorithm by evaluating different hyperparameters.
\end{abstract}


\newcommand{\cmt}[1]{}
\newcommand{\taery}[1]{\textcolor{blue}{{Taery: #1}}}
\newcommand{\sehoon}[1]{\textcolor{red}{{Sehoon: #1}}}
\newcommand{\newtext}[1]{#1}
\newcommand{\original}[1]{\textcolor{magenta}{Original: #1}}
\newcommand{\eqnref}[1]{Equation~(\ref{eq:#1})}
\newcommand{\figref}[1]{Figure~\ref{fig:#1}}
\renewcommand{\algref}[1]{Algorithm~\ref{alg:#1}}
\newcommand{\tabref}[1]{Table~\ref{tab:#1}}
\newcommand{\secref}[1]{Section~\ref{sec:#1}}

\long\def\ignorethis#1{}

\newcommand{\etal}{{\em{et~al.}\ }}
\newcommand{\eg}{e.g.\ }
\newcommand{\ie}{i.e.\ }

\newcommand{\figtodo}[1]{\framebox[0.8\columnwidth]{\rule{0pt}{1in}#1}}



\newcommand{\pdd}[3]{\ensuremath{\frac{\partial^2{#1}}{\partial{#2}\,\partial{#3}}}}

\newcommand{\mat}[1]{\ensuremath{\mathbf{#1}}}
\newcommand{\set}[1]{\ensuremath{\mathcal{#1}}}

\newcommand{\vc}[1]{\ensuremath{\mathbf{#1}}}
\newcommand{\vEndEff}{\ensuremath{\vc{d}}}
\newcommand{\vRelMove}{\ensuremath{\vc{r}}}
\newcommand{\sSet}{\ensuremath{S}}

\newcommand{\vControl}{\ensuremath{\vc{u}}}
\newcommand{\vPoint}{\ensuremath{\vc{p}}}
\newcommand{\sSpringCoef}{{\ensuremath{k_{s}}}}
\newcommand{\sDamperCoef}{{\ensuremath{k_{d}}}}
\newcommand{\vHandle}{\ensuremath{\vc{h}}}
\newcommand{\vForce}{\ensuremath{\vc{f}}}

\newcommand{\mTransChain}{\ensuremath{\vc{W}}}
\newcommand{\mRotateTrans}{\ensuremath{\vc{R}}}
\newcommand{\sJoint}{\ensuremath{q}}
\newcommand{\vJoint}{\ensuremath{\vc{q}}}
\newcommand{\mJoint}{\ensuremath{\vc{Q}}}
\newcommand{\mMass}{\ensuremath{\vc{M}}}
\newcommand{\sMass}{\ensuremath{{m}}}
\newcommand{\vGravity}{\ensuremath{\vc{g}}}
\newcommand{\vConstr}{\ensuremath{\vc{C}}}
\newcommand{\sConstr}{\ensuremath{C}}
\newcommand{\vCOM}{\ensuremath{\vc{x}}}
\newcommand{\sGeneralForce}[1]{\ensuremath{Q_{#1}}}
\newcommand{\vStateVar}{\ensuremath{\vc{y}}}
\newcommand{\vControlVar}{\ensuremath{\vc{u}}}
\newcommand{\tr}[1]{\ensuremath{\mathrm{tr}\left(#1\right)}}

%
%

\renewcommand{\choose}[2]{\ensuremath{\left(\begin{array}{c} #1 \\ #2 \end{array} \right )}}

\newcommand{\gauss}[3]{\ensuremath{\mathcal{N}(#1 | #2 ; #3)}}

\newcommand{\pctab}{\hspace{0.2in}}
\newenvironment{pseudocode} {\begin{center} \begin{minipage}{\textwidth}
                             \normalsize \vspace{-2\baselineskip} \begin{tabbing}
                             \pctab \= \pctab \= \pctab \= \pctab \=
                             \pctab \= \pctab \= \pctab \= \pctab \= \\}
                            {\end{tabbing} \vspace{-2\baselineskip}
                             \end{minipage} \end{center}}
\newenvironment{items}      {\begin{list}{$\bullet$}
                              {\setlength{\partopsep}{\parskip}
                                \setlength{\parsep}{\parskip}
                                \setlength{\topsep}{0pt}
                                \setlength{\itemsep}{0pt}
                                \settowidth{\labelwidth}{$\bullet$}
                                \setlength{\labelsep}{1ex}
                                \setlength{\leftmargin}{\labelwidth}
                                \addtolength{\leftmargin}{\labelsep}
                                }
                              }
                            {\end{list}}
\newcommand{\newfun}[3]{\noindent\vspace{0pt}\fbox{\begin{minipage}{3.3truein}\vspace{#1}~ {#3}~\vspace{12pt}\end{minipage}}\vspace{#2}}

\newcommand{\key}{\textbf}
\newcommand{\fun}{\textsc}



\newcommand{\drule}{\specialrule{0.2pt}{1pt}{1pt}%
            \specialrule{0.2pt}{0pt}{\belowrulesep}%
            }
\newcommand{\centered}[1]{\begin{tabular}{l} #1 \end{tabular}}

\newcommand{\testalgname}{Dropout-Permutation Test\xspace}

\newcommand{\hopper}{Hopper\xspace}
\newcommand{\walker}{Walker2d\xspace}
\newcommand{\halfcheetah}{HalfCheetah\xspace}
\newcommand{\pendulum}{InvertedDoublePendulum\xspace}

\newcommand{\cpupdates}[1]{\textcolor{blue}{{#1}}}
\section{INTRODUCTION}
Since the pioneering work of Mnih~\cite{mnih2015human}, deep reinforcement learning (deep RL) becomes a promising computational framework for solving high-dimensional continuous control problems in robotics.
Deep RL illustrates problems as Partially Observable Markov Decision Processes (PoMDPs) where its goal is to find optimal actions from observations.
Many deep RL algorithms have been proposed to solve various classes of PoMDPs, such as DDPG~\cite{lillicrap2015continuous}, TD3~\cite{fujimoto2018addressing}, SAC~\cite{haarnoja2018soft}, A3C~\cite{mnih2016asynchronous}, and PPO~\cite{schulman2017proximal}, just to name a few.
These algorithms are typically evaluated in simulated environments, such as OpenAI Gym~\cite{brockman2016openai} or DeepMind control suits~\cite{tassa2018deepmind}.
For a fair comparison, researchers are not allowed to change the predefined configurations of the benchmark problems.


In practice, roboticists solve their own control problems by freely defining PoMDPs that are tailored to the given robotic platforms.
One notable design challenge is to find a proper observation space, which can significantly affect the learning process.
Because there are no principled approaches or guidelines, roboticists often design an observation space solely based on prior knowledge.
A typical choice is to start with a simple concatenation of all the raw sensor inputs, such as joint positions, velocities, or inertial measurements, and further extend it with Cartesian transformations~\cite{peng2018deepmimic,wu2019mat}, binary contact flags~\cite{peng2017deeploco,tsounis2020deepgait,wu2019mat}, a short history~\cite{haarnoja2018learning,Hwangboeaau5872,yang2020data,ha2020learning}, or global transformation with state estimation~\cite{faust2018prm,Hwangboeaau5872,jeong2016bilateral}.
The problem formulation is repetitively revised based on trial-and-errors until they get satisfactory experimental results. 
Therefore, researchers often select different observation spaces for the same control problems, even for popular benchmark problems~\cite{brockman2016openai,tassa2018deepmind} (Table~\ref{tab:intro}).

\begin{table}
\begin{center}
\begin{tabular}{c|c|c}
\toprule
 & \textbf{OpenAI Gym} & \textbf{\shortstack{DeepMind \\Control Suite}} \\ 
 \drule
 \shortstack{Inverted\\DoublePendulum} 
 & \shortstack{Trigonometric \\ joint positions} 
 & \shortstack{Global \\ body rotations} \\
 \midrule
 Hopper
 & \shortstack{No \\ contact flags} 
 & \shortstack{Binary \\ contact flags} \\
 \midrule
 Walker2D
 & \shortstack{Joint positions + \\ Global root orientation} 
 & \shortstack{Global \\ body orientations} \\
\bottomrule
\end{tabular}
\end{center}
\caption{Highlights of the differences between the observation spaces from OpenAI Gym and DeepMind Control Suite benchmarks.
}
\label{tab:intro}
\vspace{-0.3cm}
\end{table}

The optimization of an observation space is connected to the design optimization of the robot: ``what is the optimal sensor configuration for the given task?’’
In other words, we want to find a sensor set that fits within the cost and space limits but guarantees the maximum performance.
For instance, a lot of legged robots~\cite{bosworth2015super,katz2019mini,unitree2019aliengo} have binary contact sensors, which are important information to develop a Raibert-style model-predictive controller~\cite{raibert1986legged}.
However, it is not clear whether a deep RL based policy can also get benefits from binary contact flags, as discovered by Reda et al.~\cite{reda2020learning}.
Therefore, it will be beneficial if we have a structured computational tool to identify the optimal sensors for the given control problem.

In this work, we first benchmark the performance of various observation spaces on the standard benchmark problems.
Our configuration includes a basic setting with joint encoders and IMUs, a reduced coordinate, a maximal coordinate, and manually designed observations, where all can be further augmented with contact flags and a short history.
Then we establish a few principles based on the experimental results.
In addition, we propose a search algorithm for automatically optimizing the observation space, which is designed based on the identified principles.
Our algorithm consists of two main components: an optimization algorithm for searching observation spaces and a statistical test for deleting malicious observation channels.
We demonstrate that our algorithm can find an optimal observation space that outperforms the manually designed observations.

Our key contributions are as follows: (1) we identify a set of principles to design observation spaces based on the thorough benchmark results, and (2) we design a novel evolutionary algorithm for automatically finding the optimal observation space, which reduces the burden of trial-and-errors.


\section{RELATED WORK}

\noindent\textbf{MDP Design Choices.}
Deep reinforcement learning has been known for its sensitive performance with respect to many design choices, hyperparameter selection, and implementation details, which make it hard to reproduce the results of prior work~\cite{islam2017reproducibility,henderson2017deep}.
Researchers have studied critical design choices for deep RL algorithms via thorough experiments, such as policy and learning configurations~\cite{andrychowicz2020matters} or code-level implementation details~\cite{engstrom2020implementation}.
On the other hand, the work of Peng and his colleague~\cite{peng2017learning} investigated MDP formulation by comparing four different action spaces: position control, velocity control, torque control, and muscle-activations.
The recent work of Reda et al.~\cite{reda2020learning} investigated various aspects of the MDP formulation, including states, observations, terminal conditions, and a few learning details.
Instead, our work conducts a more focused study on design choices for observation spaces and leverages the experimental results to develop a search algorithm.

\noindent\textbf{Network Architecture Search.}
Researchers have proposed search algorithms for finding better network architectures~\cite{angeline1994evolutionary,harp1990designing}, which are typically based on evolutionary algorithms.
One notable work is Neuro-Evolution of Augmenting Topologies (NEAT)~\cite{stanley2002evolving} that simultaneously searches network architectures and policy parameters.
Recently, a similar idea of evolutionary search has been also investigated in the context of deep learning to optimize neural network architecture~\cite{zoph2016neural,real2019regularized,miikkulainen2019evolving,chiang2019learning,gaier2019weight}, reward functions~\cite{chiang2019learning}, and other hyperparameters~\cite{miikkulainen2019evolving}.
Our problem of finding the optimal observation space is also closely related to other prior methods because it can be viewed as network architecture search with sparsity regularizations on the input layers.

\noindent\textbf{Network Pruning.}
Network pruning is a method to remove redundant connections in neural networks to avoid over-fitting, which can be also considered as the special case of network architecture search.
There have been several studies~\cite{lecun1990optimal,hassibi1993optimal,han2015learning,molchanov2017variational,zhou2019deconstructing} based on the observation that connection weights are highly correlated with their significance. 
Pruned networks are expected to perform as well as the original networks, while some studies reported even improved performances after pruning~\cite{frankle2018the,lee2018snip,liu2018rethinking,zhou2019deconstructing}.
Our methodology aligns with network pruning in terms of removing unwanted connections iteratively, during the optimization process.
Our pruning criterion is also comparable to the Neuron Importance Score Propagation (NISP)~\cite{yu2017nisp}, which measures the pruning effects on the entire network.

\noindent\textbf{Dropout for Deep RL.}
Dropout~\cite{srivastava2014dropout} is another technique widely used in deep learning to avoid overfitting. 
In deep RL, dropout is suggested to measure the uncertainty of a model based on Bayesian approximation~\cite{gal2016dropoutuncertainty,gal2016improving,gal2017concrete}, which is further leveraged to determine the switch between exploitation and exploration.
Kahn et al. estimated the uncertainty of a model using dropout with bootstrapping for collision avoidance~\cite{kahn2017collision}. 
It is also possible to use dropout during training to obtain a robust model, such as Message-dropout~\cite{kim2019messagedropout} that adopts dropout to input-level messages exchanged between multiple agents.
In this work, we also apply a dropout technique to an input layer to measure the sensitivity of the learned policy to observations.

\section{Benchmark on Observation Spaces}
\label{sec:benchmark}
This section describes benchmark experiments for comparing different sets of observation spaces.
Particularly, we design the experiments to verify commonly known design choices, such as Cartesian transformations~\cite{wu2019mat}, binary contact flags~\cite{peng2017deeploco,tsounis2020deepgait,wu2019mat}, a short history~\cite{haarnoja2018learning,Hwangboeaau5872,yang2020data,ha2020learning}, or global transformation with state estimation~\cite{faust2018prm,Hwangboeaau5872,jeong2016bilateral}.
We also aim to compare the performance of hardware-compatible configurations to common observation spaces in the RL community.

\subsection{Experimental Setup} \label{sec:benchmark_setting}

We compare various observation spaces on standard benchmark problems, InvertedDoubleCartpole-v2, Hopper-v2, Walker2d-v2, and HalfCheetah-v2, provided by OpenAI~\cite{brockman2016openai}. 
Besides the observation spaces, we do not change other MDP configurations, like action spaces, reward functions, and initial state distributions.
We select Soft-Actor Critic with Automated Entropy Adjustment (SAC-AEA)~\cite{haarnoja2018soft} as a learning algorithm, which is an off-policy deep RL algorithm with a structured exploration strategy based on the entropy constraint.
For all experiments, we set a learning rate as 0.003 and a policy architecture as a feedforward network of $[64, 64]$ for \pendulum and $[256, 256]$ for other benchmarks.
We run each experiment with ten different random seeds to reduce the variance of the results.
Please note that the benchmark results can differ from the original author’s paper~\cite{haarnoja2018soft} due to the implementation details and hyperparameters.

\subsection{Observation Spaces}
\label{sec:benchmark_observation}

\begin{table}
\begin{center}
\begin{tabular}{ |c|l| } 
 \hline
 Term & Description \\ \hline
 $\vc{q}_{rt}$ &  global position and orientation of the root. $\vc{q}_{rt} = [\vc{C}_1^T, r_1^T]^T$ \\ \hline
 $x, y, z$ & a shorthand notation of global position. $[x, y, z]^T = \vc{C}_1$\\ \hline
 $\vc{q}_{jt}$ &  joint angles. \\ \hline
 $\vc{q}$ & positions in generalized coordinate. $\vc{q} = [\vc{q}_{r}^T, \vc{q}_{jt}^T]^T$ \\ \hline
 $\vc{C}_i$ & the global position of COM of the $i$th body node. \\ \hline
 $\vc{r}_i $& the global rotation of the $i$th body in Euler angles. \\ \hline
 $\vc{R}_i $& the global rotation of the $i$th body as a Rotation matrix. \\ \hline
 $\vc{c}_i$ & a binary contact flag of the $i$th body node. \\ \hline
\end{tabular}
\end{center}
\caption{Definition of Observation Channels}
\vspace{-0.5cm}
\label{tab:terms}
\end{table}

First, we define observation channels in Table~\ref{tab:terms}. 
The defined channels may include redundant representations for the same physical quantities: for instance, the same arbitrary articulated rigid body can be represented using general coordinates or maximal coordinates. 
Some channels can be easily collected from sensors, such as $\vc{q}_{jt}$ from motor encoders or  $\dot{\vc{r}}_i$ from IMU, while some of the others cannot be easily obtained for real robots, such as the global positions $\vc{C}_1$, unless we compute those quantities via state estimation.

Based on the predefined terms, we define the following observation space configurations.
\begin{itemize}
    \item \textbf{Raw sensors (RS)}: Raw sensor readings from motor encoders and an inertial measurement unit (IMU); $\mathcal{O}^{RS} = \{\vc{q}_{jt}, \vc{\dot{q}}_{jt}, \theta, \dot{\theta}, \vc{\ddot{C}}_1 \}$. We assume that an IMU can provide a global orientation using on-board integration and filtering algorithms. 
    \item \textbf{Generalized coordinates (GC)}: Positions and velocities in generalized coordinates; $\mathcal{O}^{GC} = \{\vc{q}, \vc{\dot{q}}\}$.
    \item \textbf{Maximal coordinates (MC)}: Positions and velocities in Cartesian coordinates. Both positional and rotational velocities are included; $\mathcal{O}^{MC} = \{\vc{C}_{1\cdots N},\vc{\dot{C}}_{1\cdots N}, \vc{r}_{1\cdots N}\}$.
    \item \textbf{Open AI (OAI)}: Default observation space suggested in OpenAI benchmarks; $\mathcal{O}^{OAI} = \{z, \theta, \vc{q}_{jt}, \vc{\dot{q}} \}$
\end{itemize}
For all configurations, we subtract the global horizontal position $x$ to the global positions $\vc{C}_i$ or $\vc{q}_{rt}$ to make the observation space invariant to horizontal translations.

In addition, we extend the observation space of the RS configuration with a global height, contact flags, and maximal coordinates to investigate their effects.
\begin{itemize}
    \item \textbf{RS with contact flags (RS+C)}: $\mathcal{O}^{RS} \cup \{\vc{c}_{1 \cdots K} \}$, where $K$ is the contact sensors. We attach contact sensors to the tip of each foot of \hopper and \halfcheetah, and the toe and heel of \walker agent. There are no contact sensors for \pendulum.
    \item \textbf{RS with positions in Cartesian coordinates (RS+CP)}: $\mathcal{O}^{RS} \cup \{\vc{C}_{1\cdots N}\}$.
\end{itemize}
We finally propose a new observation space by leveraging the prior knowledge and benchmark results:
\begin{itemize}
    \item \textbf{Ours}: $\mathcal{O}^{RS} \cup \{z,\dot{\vc{C}_1}\}$.
\end{itemize}
We also extend RS, OAI, and Ours with a short $N$ history of actions, by concatenating $N$ previous observation histories $\vc{o}_t, \vc{o}_{t-1}, \cdots, \vc{o}_{t-N+1}$ and $N-1$ previous actions $\vc{a}_{t-1}, \cdots, \vc{o}_{t-N+1}$.
\subsection{Benchmark Results and Analysis}

\begin{figure}
    \centering
    \includegraphics[width=1\linewidth]{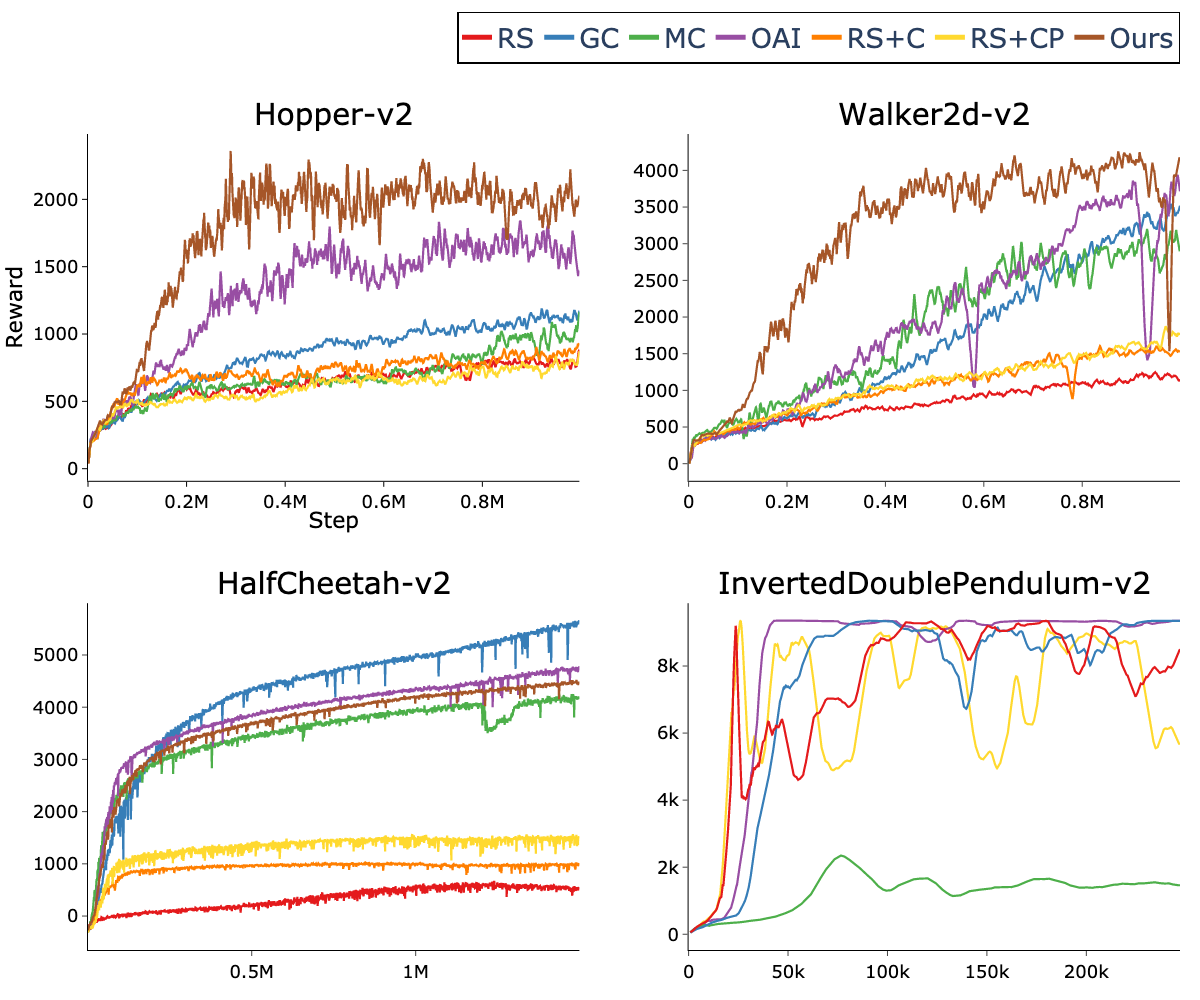}
\caption{
Learning curves with different observation set. Benchmark environments: Hopper, Walker2D, HalfCheetah, InvertedDoublePendulum.
}
    \label{fig:b1-obs_set}
\end{figure}

\begin{figure}
    \centering
    \begin{tabular}{c c}
        \includegraphics[width=0.47\linewidth]{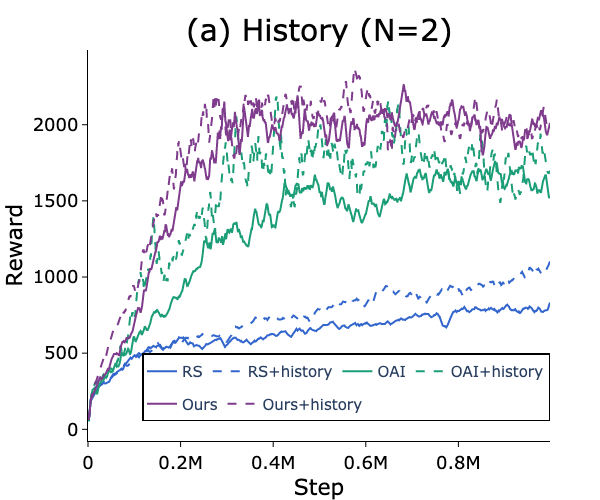}
        \includegraphics[width=0.47\linewidth]{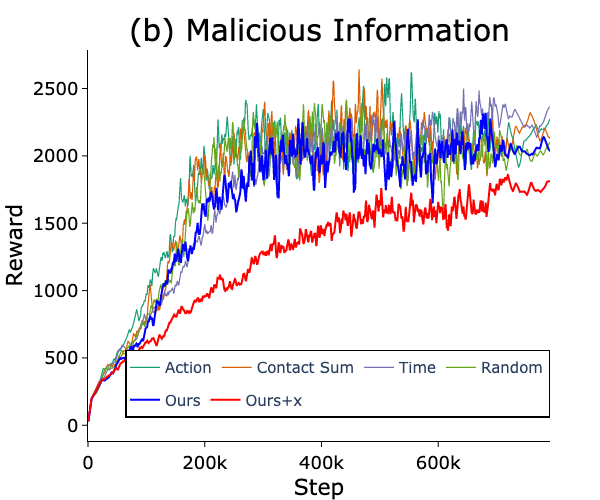}
    \end{tabular}
    \vspace{-1em}
    \caption{
    Learning curves on Hopper-v2.
    (a) Three observation spaces (RS, OAI, and Ours) with and without a short history.
    (b) Ours set with different additional information, which drops significantly when we add $x$ to Ours.
    }
    \label{fig:b2-malicious}
\end{figure}

\noindent \textbf{General Comparison.}
Figure~\ref{fig:b1-obs_set} illustrates the comparison of various observation spaces: RS, GC, MC, OAI, RS+C, RS+CP, and Ours.
In general, all configurations learn reasonably well on the \pendulum environment.
For the other three locomotion tasks, a ``reactive'' environment with only the RS observations shows the worst learning progress on average.
However, augmenting the RS with a global height and a linear velocity (Ours) has a notable positive impact, which demonstrates the importance of global state estimation for locomotion.
We do not observe any notable difference between the learning curves of GC and MC which indicates that the choice of coordinates does not have a significant impact on the results. Similar to the recent study of Reda et al.~\cite{reda2020learning}, we see that adding positions in Cartesian coordinates can be helpful in complex environment by comparing RS and RS+CP on the \halfcheetah environment. 
The default settings of OpenAI Gyms (OAI) result in reasonably efficient learning, although a few settings outperform the default observation spaces.

\noindent \textbf{Contact Information.}
One commonly known fact in robotics is that contact flags are crucial for developing an effective locomotion controller, particularly for a Raibert-style MPC controller~\cite{raibert1986legged}.
To this end, we investigate the impact of contact flags by augmenting the raw sensor configuration (RS) with discrete contact flags (RS+C) and positions in Cartesian coordinates (RS+CP).
In the RS+CP configuration, the position of each body contains the distance to the flat terrain, which can be viewed as a continuous version of contact information.
However, we observe minor performance differences for all three observation spaces, except minor performance gains on the \halfcheetah environment.
This indicates that RL-based locomotion policies do not particularly require explicit contact information, which is well-matched with the conclusion of the study~\cite{reda2020learning} done on PyBullet locomotion benchmarks~\cite{coumans2019}.

\noindent\textbf{History of Actions and States.}
There exist a few prior works that augment the observation space with a history of previous observations and actions for more efficient learning and a more robust final policy.
However, it can degrade the learning speed due to the increased dimension of the observation space.
We test this hypothesis by augmenting the configuration RS, OAI, Ours with an N-history of states and actions, where N = 2 for all the experiments.
The benchmark results are illustrated in Figure~\ref{fig:b2-malicious}a.
The results indicate that an N-history is slightly helpful for learning, particularly in the case of RS, which lacks processed information. 

\noindent\textbf{Malicious Information.}
It is a well-known fact that it is better to exclude some task-irrelevant information from the observation space, such as the horizontal root position $x$ for locomotion tasks, which can be malicious for the entire learning process, which was verified with our experiments (Fig.~\ref{fig:b2-malicious}b, Ours+x).
However, it is not intuitive to find another malicious observation channel that significantly slows down the learning process.
We investigate other candidates, such as time variable $t$, a frame counter, previous action, cumulative sum of contact flag, or random numbers, but none of them showed noticeable impacts on the learning curve~(Fig.~\ref{fig:b2-malicious}b).
We suspect that a malicious sensory channel must be deceptive enough to 1) dominate the initial learning process but 2) not relevant for longer horizons.







\section{Optimization of Observation Spaces}

In the previous sections, we benchmark different observation spaces in OpenAI Gym environments.
The results indicate that the selection of observation spaces has notable impacts on the learning process, and different problems have their own optimal observation spaces.
However, it is not possible to simply determine whether one observation channel is helpful to the given problem or not because learning speed depends on the combination of observation channels, not a single channel.
Also, a single malicious information channel can slow down the learning significantly.
This motivates an algorithmic approach that can automatically find the optimal observation space to the given problem.

\subsection{Search Algorithm}

\begin{algorithm}
	\caption{Optimization of Observation Spaces}
	\textbf{Inputs:} Model, observation, training time $K$, test\_metric
	\label{alg:randomsearch}
	\begin{algorithmic}[1]
        \State initial\_obs ← basic sensors
        \State basic\_model ← make\_model(initial\_obs)
        \State basic\_model.learn($K$)
        \State basic\_score ← basic\_model.test(test\_metric)
        \State best\_score, best\_obs ← basic\_score, basic\_obs
        \While{not converged}
            \State current\_obs ← best\_obs $\cup$ new\_obs\_group
            \State current\_model ← make\_model(current\_obs)
            \State current\_model.learn($K$)
            \State current\_score ← current\_model.test(test\_metric)
            \If {current\_score $\>$ best\_score}
                \State current\_obs ← remove\_malicious(current\_obs)
                \Comment{We use Algorithm~\ref{alg:permutation}}
                \State best\_score, best\_obs ← current\_score, current\_obs
            \EndIf
        \EndWhile
	\end{algorithmic} 
\end{algorithm}

We propose an optimization algorithm (Algorithm~\ref{alg:randomsearch}) to find the most effective observation space for the given MDP problem.
We start from the initial observation space, which is set to the raw sensor (RS) configuration, and gradually investigate different observation spaces.
At each iteration, we change the observation space by adding random observation channels.
Then we train the model for $K$ steps 
and evaluate the performance of the updated observation space.
For evaluation, we can generate additional rollouts or take an average of the episodic rewards from previous training rollouts.
We accept this new observation space only if its performance is improved.
Different metrics to evaluate the performance will be examined in Section~\ref{sec:ablation}.

One important design factor of our algorithm is how to mutate the current observation space.
In our experience, it is better to take more \emph{add} operations rather than \emph{delete} operations.
This is also well matched our benchmark results, which show that the redundancy is quite acceptable in many situations.
To this end, our algorithm proposes a new observation space by adding a group of observations, without \emph{delete} operations.
We select a group based on their semantics, instead of selecting random observation channels.
We empirically found that this semantic search yields the best results, which will be further studied in Section~\ref{sec:ablation}.

However, this algorithm does not produce a concise observation space because it cannot delete observation channels once it accepts them.
Therefore, we allow our algorithm to delete observation channels via statistical tests, which will be explained in the following section.

\subsection{Statistical Test for Removing Unnecessary Channels} \label{sec:dropout}

\begin{algorithm}
	\caption{\testalgname for removing unnecessary channels.}
	\label{alg:permutation}
	\textbf{Inputs:} observation\_space, dropout rate $d$, threshold $\bar{I}$
	\begin{algorithmic}[1]
	    \State model ← make\_model(observation\_space)
	    \State model.learn(n\_steps, dropout=(model.input\_layer, $d$))
	    \State base\_score ← model.test()
		\For {$i=1,2,\ldots,N$}
			\State score ← model.test(obs$_i$=sample(range=obs\_range))
    	    \State importance ← (score$-$base\_score) / base\_score
		\EndFor
		\State remove observation channels where its importance $< \bar{I}$
		\State \Return observation\_space
	\end{algorithmic} 
\end{algorithm}

We propose a statistical test, which is the so-called \emph{Dropout-Permutation Test}, to identify the importance of each channel in the given observation space.
First, we train an additional robust policy by applying a dropout technique to the input layer.
This prevents our algorithm from being too sensitive to each observation channel.
Then we iteratively examine each observation channel by substituting the input with a randomized value sampled from the range of each channel.
If the performance drops significantly, it means that the randomized channel was important to train the given policy.
However, if the performance is not changed or even increased, it indicates that the channel is not necessary or even malicious.
Our algorithm is summarized in Algorithm~\ref{alg:permutation}.

Our test is inspired by perturbation-based saliency methods, which have been applied to explain the saliency of the actions.
Depending on the environments, prior studies perturb by masking pixels in an image~\cite{zeiler2014visualizing,greydanus2018visualizing,iyer2018transparency} or removing pieces on a game board~\cite{puri2020explain}.
Here, we substitute with a randomized value sampled from a range of input data collected during learning to see how much an agent gets confused by a fake but realistic input. 
\section{Results}
\label{sec:experiments}

In this section, we discuss the experiments for validating the proposed optimization algorithm.
Particularly, we design the experiments to answer the following questions.
\begin{enumerate}
    \item Can the Dropout-Permutation test find unnecessary or malicious information channels?
    \item Can our algorithm find an optimal observation space that is better than the manually designed ones?
    \item How does the selection of hyper-parameters impact the performance of the algorithm?
\end{enumerate}

\subsection{Experimental Setup}
We also evaluate our algorithm on four MuJoCo~\cite{todorov2012mujoco} environments in OpenAI Gym~\cite{brockman2016openai}, with the same hyper-parameters for the environments and learning algorithm as described in Section~\ref{sec:benchmark_setting}.
For the optimization algorithm, we initialize the sensor set as Raw Sensors (RS), use the group-based selection scheme, set the number of pre-training timesteps as 200k (30k for \pendulum), and set the evaluation scheme as \textit{All}.
The semantic search space includes $\vc{C}_1$, $\dot{\vc{C}_1}$, Cartesian coordinates, contact flags, and previous action. 
The effects of these hyper-parameters will be analyzed in Section~\ref{sec:ablation}.
We run each experiment with ten different random seeds, which is the same as our benchmark experiments.

\subsection{Dropout Test} \label{sec:exp-dropout}
We conduct experiments to validate whether our \testalgname test can successfully measure the importance of each observation channel.
We apply the test to the Ours+x observation space on the Hopper-v2 environment, which contains additional malicious information of the horizontal position $x$.
Also, we carry out permutation test on the Ours set for a comparison. 
Intuitively, if the performance drops upon the removal of one observation channel, it implies that the channel contains relevant information to the task.
If the performance does not drop or even increases, the observation might not be important or even malicious to the learning.

\begin{figure}
    \centering
    \includegraphics[width=1\linewidth]{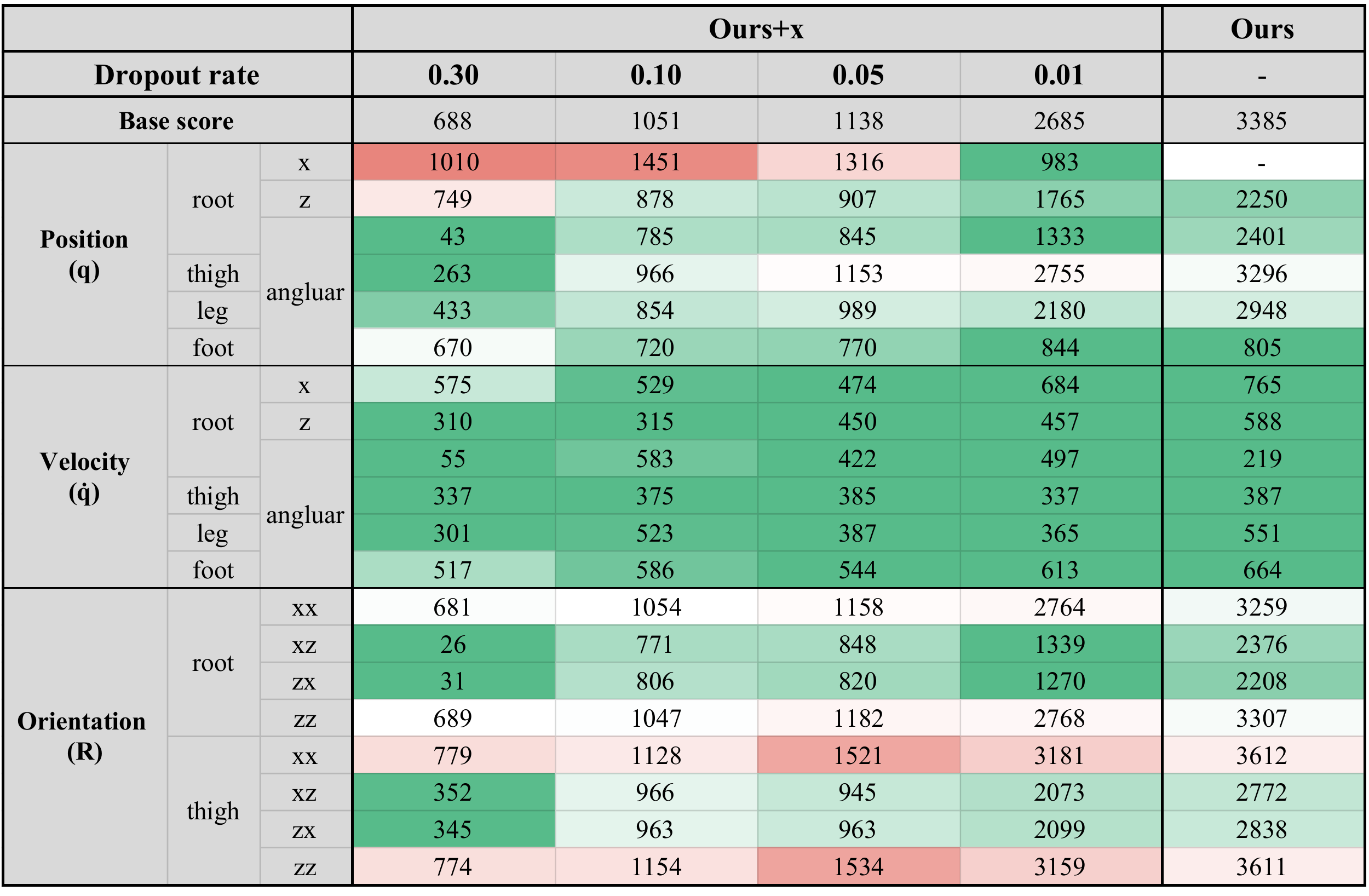}
    \vspace{-2em}
    \caption{
    Visualization of \testalgname result on Hopper-v2 with respect to dropout rate on \textit{Ours+x}, compared with the permutation score of \textit{Ours} (condensed). 
    Red: increased test scores, malicious; Green: decreased test scores, essential.
    When dropout rate is low, we can expect higher average test rewards, but it becomes difficult to detect malicious information.
    }
    \label{fig:e1-dropout}
\end{figure}

Figure~\ref{fig:e1-dropout} shows the visualization of the \testalgname results, with different dropout rates of [0.3, 0.1, 0.05, and 0.01].
Green cells indicate useful information channels while red cells are not relevant or malicious channels.
In general, a higher dropout rate makes the policy less effective: the average reward with the dropout rate 0.3 is 688.2, which is much smaller than the average score 2685.2 of the dropout rate 0.01.
However, it is harder to identify the importance of observation channels, by being less robust to the removal of observations.
Therefore, we choose a dropout rate as 0.1 for our experiments.

\subsection{Search Algorithm} \label{sec:result-search}

\begin{figure}
    \centering
    \includegraphics[width=1\linewidth]{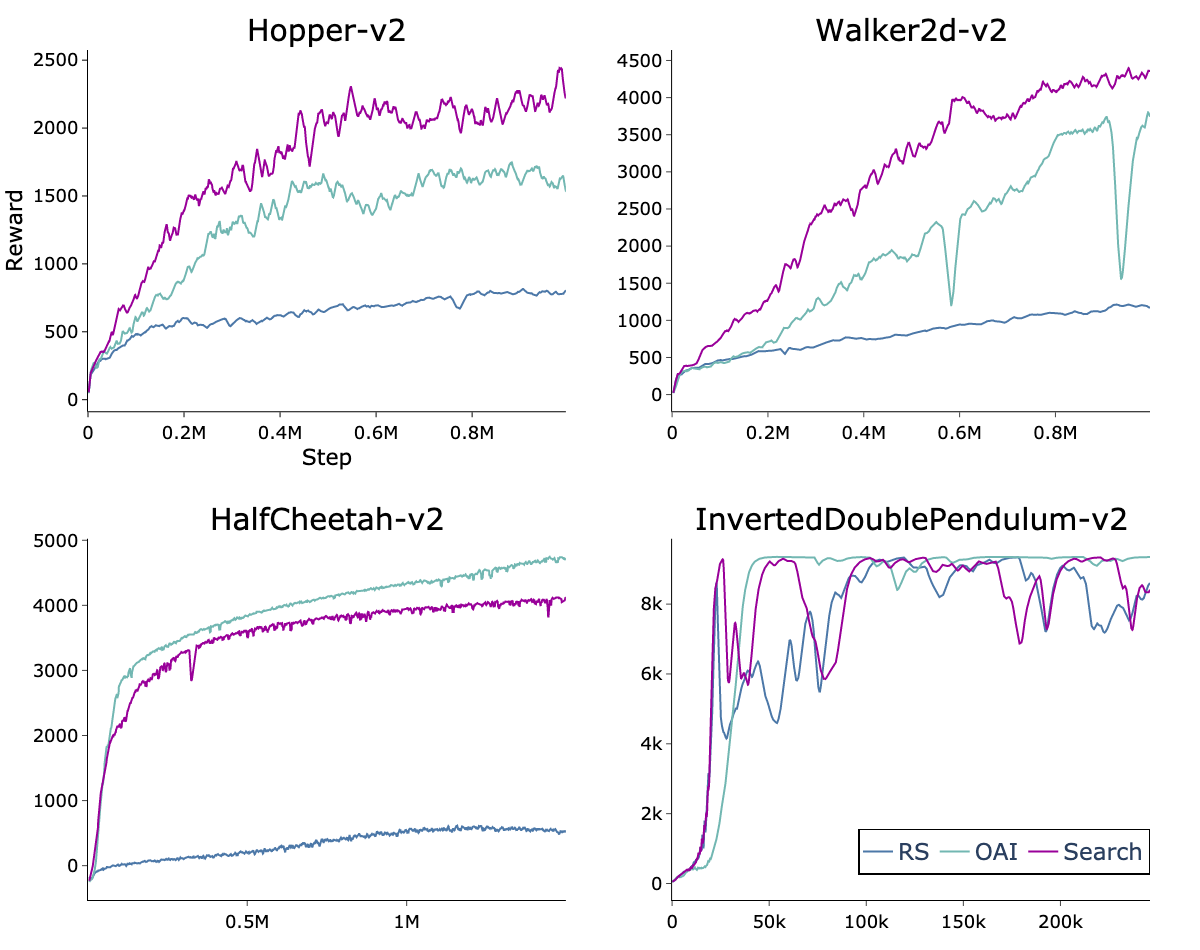}
    \vspace{-1.5em}
    \caption{Learning curves comparing optimized observation set (Search) to initial observation set (RS) and OpenAI benchmarks observation (OAI).}
    \label{fig:exp-2-result}
\end{figure}

\begin{table}
\begin{center}
\begin{tabular}{c|c|c|c|c|c}
\toprule
 & \textbf{$\vc{C}_1$} & \textbf{$\dot{\vc{C}_1}$} & \textbf{\shortstack{Cartesian\\coordinates}} & \textbf{Contact} & \textbf{\shortstack{Prev.\\action}}\\ 
\drule
Hopper  & 1 & 9 & 4 & 0 & 6\\ 
\midrule 
Walker2D  & 1 & 10 & 7 & 3 & 6\\ 
\midrule 
\shortstack{HalfCheetah}  & 4 & 7 & 4 & 2 & 4\\ 
\midrule 
\shortstack{Inverted\\DoublePendulum}  & 0  & 1  & 4  & -  & 0\\
\bottomrule
\end{tabular}
\end{center}
\vspace{-0.5em}
\caption{Optimized result.
The number indicates the count of runs among ten different runs that selected the given group.
}
\vspace{-0.8em}
\label{tab:selected}
\end{table}

We evaluate the algorithm on the benchmark problems.
We compare the optimized observation space against two baselines: Raw Sensors (RS) and OpenAI Gym (OAI).
RS is the initial observation space of our search algorithm, so the comparison with the RS will highlight the performance gain after the optimization.
And the comparison with OAI will show the benefit of optimization over manual approaches.

Figure~\ref{fig:exp-2-result} illustrates the learning curves of RS, OAI, and Ours on the benchmark problems.
Overall, our algorithm finds effective observation spaces that lead to better or comparable learning speeds than both RS and OAI.
However, our algorithm fails to find the optimal observation space for \halfcheetah.
We suspect that this is due to the special characteristic of \halfcheetah.
It is less sensitive to a balancing issue compared to the other two locomotion tasks, so it works better with a minimal observation space, such as Generalized Coordinates (GC), as represented in Figure~\ref{fig:b1-obs_set}.
However, we tune our algorithm to have a high chance to include redundant observation channels, which is not effective for this problem.

Table \ref{tab:selected} shows how often each semantic group has selected by our algorithm from ten different runs. 
Therefore, we can infer that the sensor groups with more occurrences are important than the groups with fewer occurrences.
For locomotion tasks, most of the runs pick $\dot{\vc{C}}_1$, which shows the importance of the linear velocity for maintaining the balance.
Cartesian coordinates is also selected quite often, but a bit more for the \walker.
Contact information is often known to be important for robotic locomotion, particularly for highly dynamic motions. 
However, it is not frequently selected for locomotion tasks, particularly zero times for \hopper.
In the case of \pendulum, RS already shows the near-optimal performance while Cartesian coordinates is added for four out of ten trials.
The results align well with our benchmark results in Section~\ref{sec:benchmark}.


\subsection{Analysis} \label{sec:ablation}
In this section, we demonstrate how design choices in our algorithm make differences on learning.  
For all experiments, we select \hopper as the testing environment.

\begin{figure}
    \centering
    \begin{tabular}{c c}
        \includegraphics[width=0.47\linewidth]{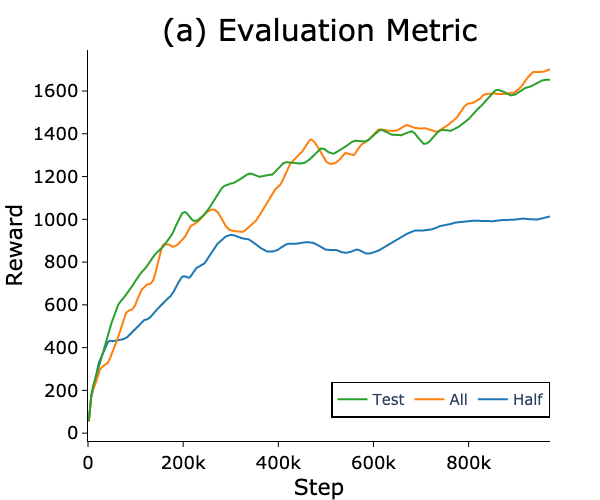}
        \includegraphics[width=0.47\linewidth]{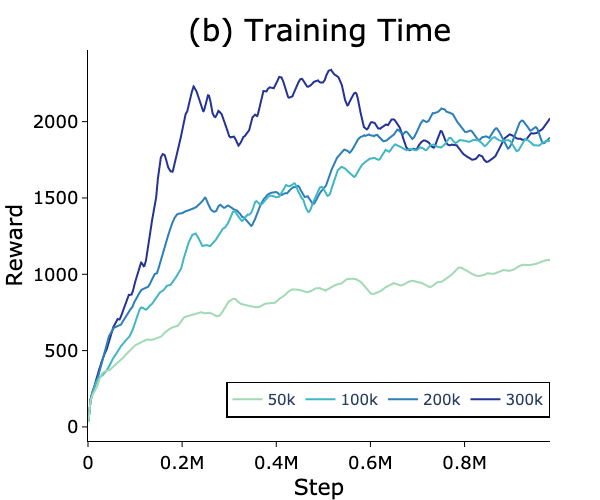} \\
        \includegraphics[width=0.47\linewidth]{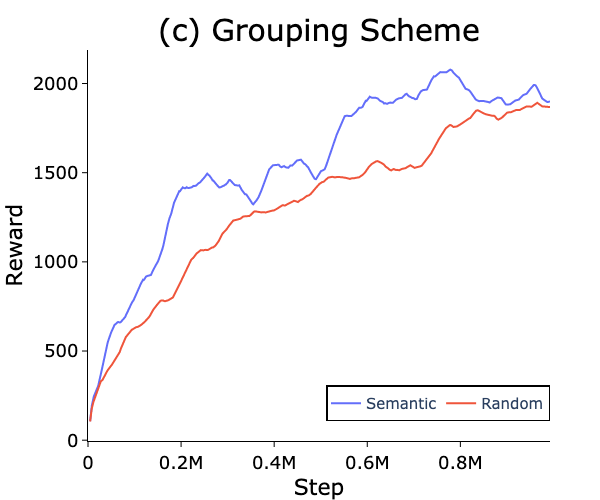}
        \includegraphics[width=0.47\linewidth]{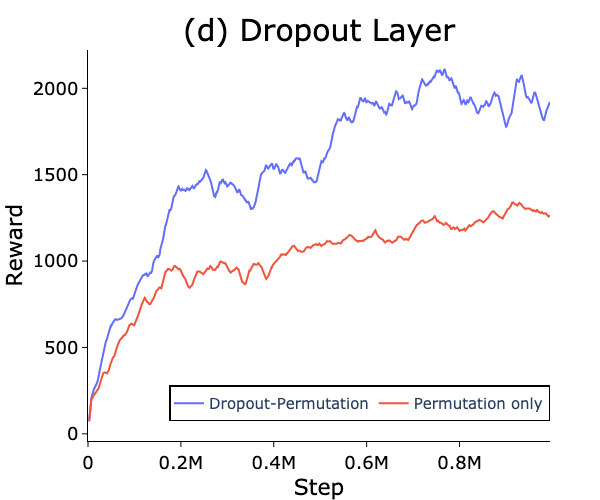}        
    \end{tabular}
    \caption{
    Comparison of design choices. The plots show learning curves of the optimized observation spaces. We evaluate four different hyperparameters, (a) evaluation metric, (b) training time, (c) grouping scheme, and (d) dropout layer.
    }
    \label{fig:exp-abl-1}
\end{figure}

\noindent\textbf{Evaluation metric.}
There can be several methods to assess the performance of the observation space in each iteration. We compare three different methods, (a) testing the trained model by generating additional rollouts (\textit{Test}) or measuring the history of rewards during (b) the entire training time (\textit{All}) or (c) later half of the training (\textit{Half}). 
Figure~\ref{fig:exp-abl-1}a illustrates the learning curves optimized with different evaluation metrics on \hopper.
In our experiments, \textit{All} and \textit{Test} show comparable performances while \textit{Half} shows the saturated learning curves.
We choose to use \textit{All} for all the experiments because it does not require extra computation costs compared to \textit{Test}.

\noindent\textbf{Training time $K$ for each iteration.}
For each iteration in Algorithm~\ref{alg:randomsearch}, we train a model for K steps prior to evaluation.
Intuitively, training more steps leads to better estimation of the final learning but requires extra costs.
As shown in Figure~\ref{fig:exp-abl-1}b, the learning speed significantly drops when training duration is less than 100k steps.

\noindent\textbf{Grouping scheme.}
Another design choice to make in Algorithm~\ref{alg:randomsearch} is how to group the observations: we can group the observation semantically, as grouped in Table~\ref{tab:terms}, or randomly.
Our intuition is that observations in the same semantic groups are related and provide more useful information when combined.
Figure~\ref{fig:exp-abl-1}c shows that semantic groups result in slightly more effective learning than random groups.


\noindent\textbf{Dropout layer.}
Figure~\ref{fig:exp-abl-1}d shows the impact of dropout layer in Dropout-Permutation Test on learning curve.
When dropout is not applied, the search algorithm learns suboptimal optimization spaces, which shows the importance of dropout layers.
\section{CONCLUSIONS}


In this work, we investigated different observation spaces and developed an algorithm to optimize the observation space based on the experimental results.
Our experimental results show that a few common design choices, such as contact flags, are not beneficial to the learning speed while identifying a few important observation channels, such as the height of the robot or a history of actions. 
Then we proposed a search algorithm to optimize the observation space, which leverages a Dropout-Permutation test to remove unnecessary or malicious channels.
We demonstrated that our algorithm significantly improves learning speeds by finding the optimal observation space for the given problem.

Although we investigated benchmark problems, the performance of deep RL algorithms highly depends on problems, algorithms, and hyperparameters.
Therefore, our benchmark results may not be well generalized to new problems.
Our work supplements the study of Reda et al.~\cite{reda2020learning} that investigates many aspects of deep RL, but both studies can be specific to the examined environments.
Our optimization algorithm is simple and effective, but it requires extra costs to repetitively investigate various observation candidates.
A possible direction for future work would be to extend this work by jointly optimizing the observation spaces with the network parameters.




\bibliographystyle{IEEEtran}
\bibliography{annot}

\end{document}